# Monkeypox disease recognition model based on improved SE-InceptionV3

Junzhuo Chen[a,*], Zonghan Lu[b] and Shitong Kang[a]
[a]*School of Artificial Intelligence, Hebei University of Technology, 300130, Tianjin, China*
[b]*School of Electrical Engineering, Hebei University of Technology, 300130, Tianjin, China*

**Abstract.** In the wake of the global spread of monkeypox, accurate disease recognition has become crucial. This study introduces an improved SE-InceptionV3 model, embedding the SENet module and incorporating L2 regularization into the InceptionV3 framework to enhance monkeypox disease detection. Utilizing the Kaggle monkeypox dataset, which includes images of monkeypox and similar skin conditions, our model demonstrates a noteworthy accuracy of 96.71% on the test set, outperforming conventional methods and deep learning models. The SENet module's channel attention mechanism significantly elevates feature representation, while L2 regularization ensures robust generalization. Extensive experiments validate the model's superiority in precision, recall, and F1 score, highlighting its effectiveness in differentiating monkeypox lesions in diverse and complex cases. The study not only provides insights into the application of advanced CNN architectures in medical diagnostics but also opens avenues for further research in model optimization and hyperparameter tuning for enhanced disease recognition. https://github.com/jzc777/SE-inceptionV3-L2

## 1. Introduction

When a rare disease begins to spread rapidly across the globe, medical and public health experts are compelled to take urgent action. Since the first discovery of the monkeypox virus in 1958 at a research center in Denmark [28], this disease has been the focus of scientific attention. Monkeypox virus belongs to the same family as smallpox virus but has very different characteristics in terms of pathogenicity and mode of transmission. In recent years, attention to this rare disease has increased with the emergence of monkeypox virus cases in Africa and previously unrecorded areas. The common symptom of monkeypox is the appearance of a rash or mucous membrane lesions that can last from two to four weeks, accompanied by fever, headache, muscle aches, back pain, lack of energy, and swollen lymph nodes. Its early symptoms resemble influenza and the subsequent rash has become one of the distinguishing features of the disease [29].

Currently, we are facing an entirely new challenge: according to the World Health Organization, as of 13 December 2023, the cumulative number of monkeypox cases worldwide reached 92,976 confirmed cases and 172 cumulative deaths. Among them, the epidemic is severe in the United States, Brazil, and Spain [31]. There is an urgent need to accurately identify people infected with this highly contagious disease that resembles other symptoms. The challenge lies not only in treatment but also in timely and accurate diagnosis and differentiation of monkeypox cases. There is currently no antiviral therapy capable of curing monkeypox [30], which has prompted an urgent need for researchers to find viable methods to identify monkeypox disease and initiate data collection and research trials.

The initial diagnosis of monkeypox disease usually relies on the experience and expertise of physicians by observing the patient's skin lesions, medical history, and other characteristics. However, this approach is more subjective and requires an experienced healthcare professional to make an accurate judgment. With the development of computer vision and artificial intelligence technologies, new possibilities for monkeypox disease recognition have emerged. Using machine learning algorithms and deep learning models, by analyzing patient skin images, we can more accurately diagnose monkeypox disease and provide decision support tools for doctors.

*Corresponding author. E-mail: jzchen7@foxmail.com

Significant progress has been made in the application of deep learning models in the field of automated identification and diagnosis of monkeypox disease.2022 Shams Nafisa Ali and his team at the University of Engineering and Technology, Bangladesh, applied a deep learning model on the Monkeypox Skin Damage Dataset (MSLD). Their study found that the ResNet50 model performed best on this dataset, achieving an accuracy of 82.96% (± 4.57%), which was slightly higher than VGG16's 81.48% ( ± 6.87%) and InceptionV3's 74.07% ( ± 3.78%) [14]. This study reveals the potential of deep learning in the field of monkeypox diagnosis. Following this, in 2023, Malathi Velu and his team proposed a Q-learning-based monkeypox disease recognition method with 95% accuracy using the monkeypox disease dataset provided by the Kaggle platform [4, 5] [15]. This result reconfirmed the effectiveness of deep learning techniques in medical image processing. In the same year, Amir Sorayaie Azar et al. achieved 95.18% accuracy on a self-constructed dataset containing four different skin conditions using the improved DenseNet-201 model [25]. This illustrates the significant performance of deep learning models in more complex classification tasks through specific improvements and optimizations. Furthermore, in 2023, Entesar Hamed I. Eliwa's team combined a CNN model with the GWO optimizer to achieve 95.3% accuracy [26]. This finding points to the fact that the selection of an appropriate optimizer can enhance the model's ability to discriminate between positive and negative categories, thus improving the overall model performance. Meanwhile, Shams Nafisa Ali et al. enumerated State-of-the-art deep learning models and found that DenseNet121-TL achieved the best combined accuracy of 83.59 ± 2.11% on the HAM10000 dataset [5].

In a previous study, in 2022, Mamun and his team found that Inception-V3 showed good effectiveness in recognizing the five most common skin diseases [33]. Subsequently, in 2023, Sharma et al. argued that Inception V3-based models have great advantages in solving the recognition of skin diseases [34].InceptionV3 has won wide recognition in the field of medical image processing for the depth and efficiency of its network architecture and especially shows unique advantages in dealing with medical images with high variability and complexity.

Therefore, in this study, we propose an improved InceptionV3-based monkeypox disease recognition model. However, the traditional InceptionV3 model may not perform well enough in dealing with monkeypox lesions with high variability and complexity. Therefore, we chose InceptionV3 as the base model and improved it to better accommodate the specificity of monkeypox lesions.

To further improve the performance of the monkeypox disease recognition model and overcome the generalization and fitting problems in monkeypox disease recognition, we propose an improved InceptionV3 model. We embed the SENet module into the InceptionV3 model to enhance the model's ability to recognize monkeypox disease-specific features. In addition, to alleviate the overfitting problem, our model introduces L2 regularisation to make the model more robust in dealing with monkeypox lesions. We trained and tested our model extensively using the publicly available monkeypox disease dataset on the Kaggle platform. Experimental results show that our model exhibits more robust performance and accuracy compared to similar models. This improved InceptionV3-based algorithm is the core contribution of this study, providing a more reliable solution for monkeypox disease diagnosis.

This paper is divided into the following three main sections: first, the network structure of InceptionV3 and the related algorithm design are introduced; then, the structure of the experimental data is described, ablation and comparison experiments are conducted and the results are analyzed in detail; finally, the concluding remarks and summary of the paper are presented.

## 2. Model Architecture

### *2.1. InceptionV3 model*

The GoogLeNet network, introduced by Google in 2014, incorporates the Inception network structure [2,18] as its fundamental architecture. This design choice achieves a dual purpose of reducing the network's parameter count while increasing its depth.

Simultaneously, the model efficiently decreases the dimensions of feature maps, allowing it to preserve image features effectively while minimizing computational complexity. As a result, GoogLeNet has gained widespread popularity in various image classification tasks.

The Inception module typically consists of three convolutional layers of different sizes and a maximum pooling layer. This module is designed to process the output of the network from the previous layer, perform multi-scale feature extraction on the input data through convolutional operations, and then perform channel aggregation of these features of different scales, and finally perform nonlinear fusion. This design helps to improve the expressiveness of the network and enhances the network's ability to adapt to different scales of information, while effectively preventing overfitting. As shown in Fig 1, the Inception network structure demonstrates the composition and working principle of this module. The flexibility and performance advantages of this design make the Inception module an important component in convolutional neural networks.

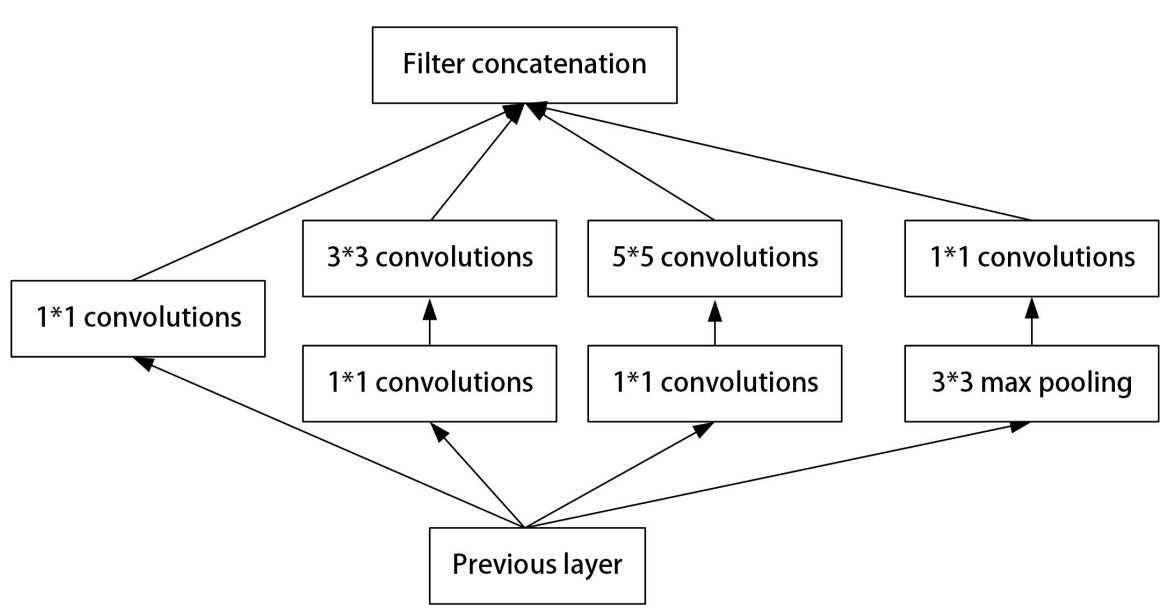

Fig. 1. Inception network structure.

InceptionV3 is a version of GoogLeNet proposed in 2015 that was pre-trained in ImageNet. The model has a default image input size of 299x299 pixels with three channels. Unlike traditional convolutional neural networks, the InceptionV3 model employs a unique method to decompose the original two-dimensional convolutional layer into two one-dimensional convolutional layers, a strategy that not only helps reduce the number of parameters but also effectively mitigates the risk of overfitting. In this paper, we adopt exactly the Inception v3 network structure, as shown in Fig 2.

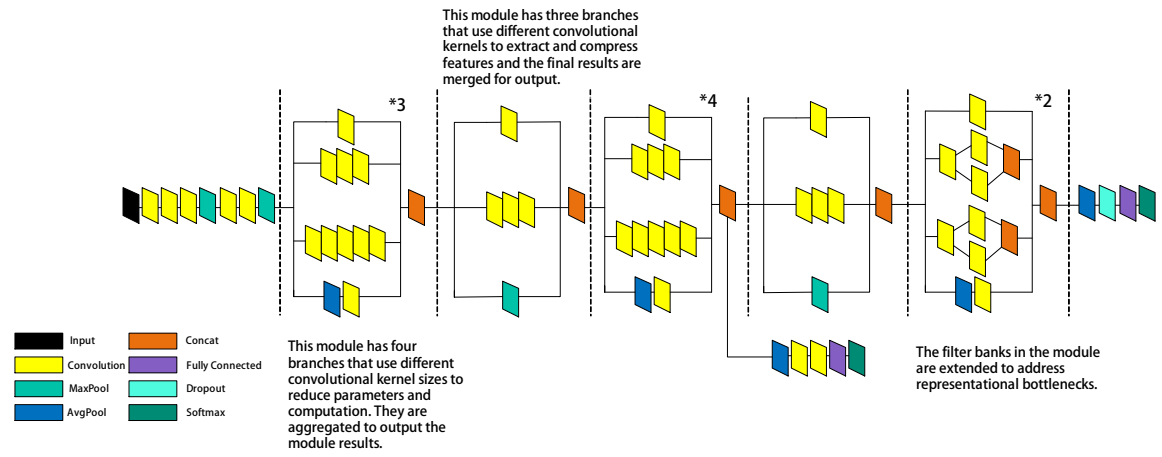

Fig. 2. Inception v3 network structure.

Compared to the previous versions InceptionV1 and V2, the network architecture of Inceptionv3 employs a convolutional kernel splitting strategy, where a larger convolutional kernel is split into multiple smaller convolutional kernels.

For example, splitting a $5 \times 5$ convolutional operation into two independent $3 \times 3$ convolutional operations is an effective strategy in order to increase computational speed and reduce the number of parameters. In fact, a $5 \times 5$ convolution operation is equivalent to 2.78 times two $3 \times 3$ convolution operations in terms of computational cost. Therefore, by splitting it into two $3 \times 3$ convolutional operations, we can realize a significant improvement in performance. This splitting method not only helps to speed up the training of the network but also captures the spatial features of the image more effectively. This is shown in the left panel of Fig 3.

In summary, InceptionV3 optimizes the Inception network structure module by introducing three different sizes of region grids, as shown in Fig 3. This improvement makes the network more adaptable to features of different sizes and improves the performance and generalization of the model.

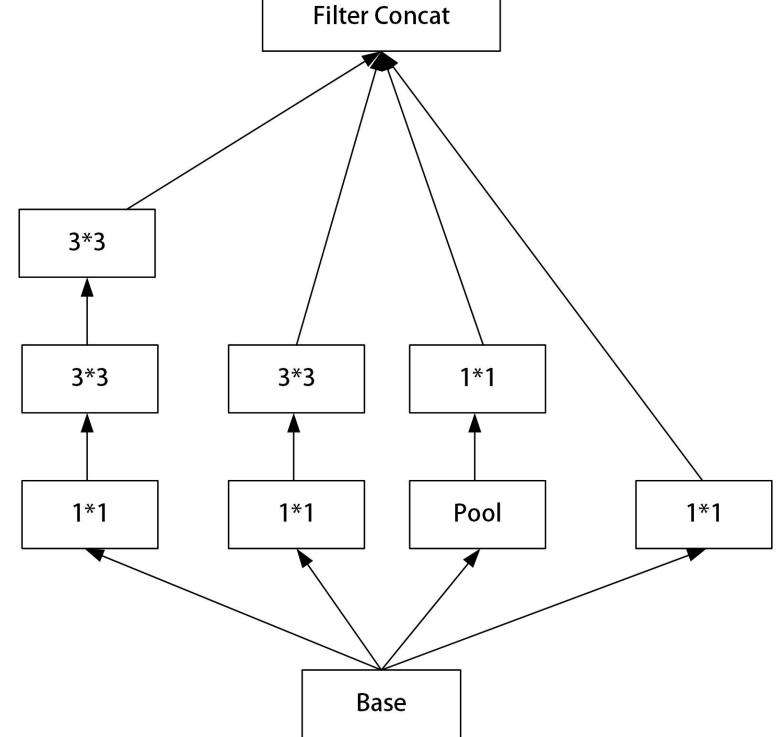

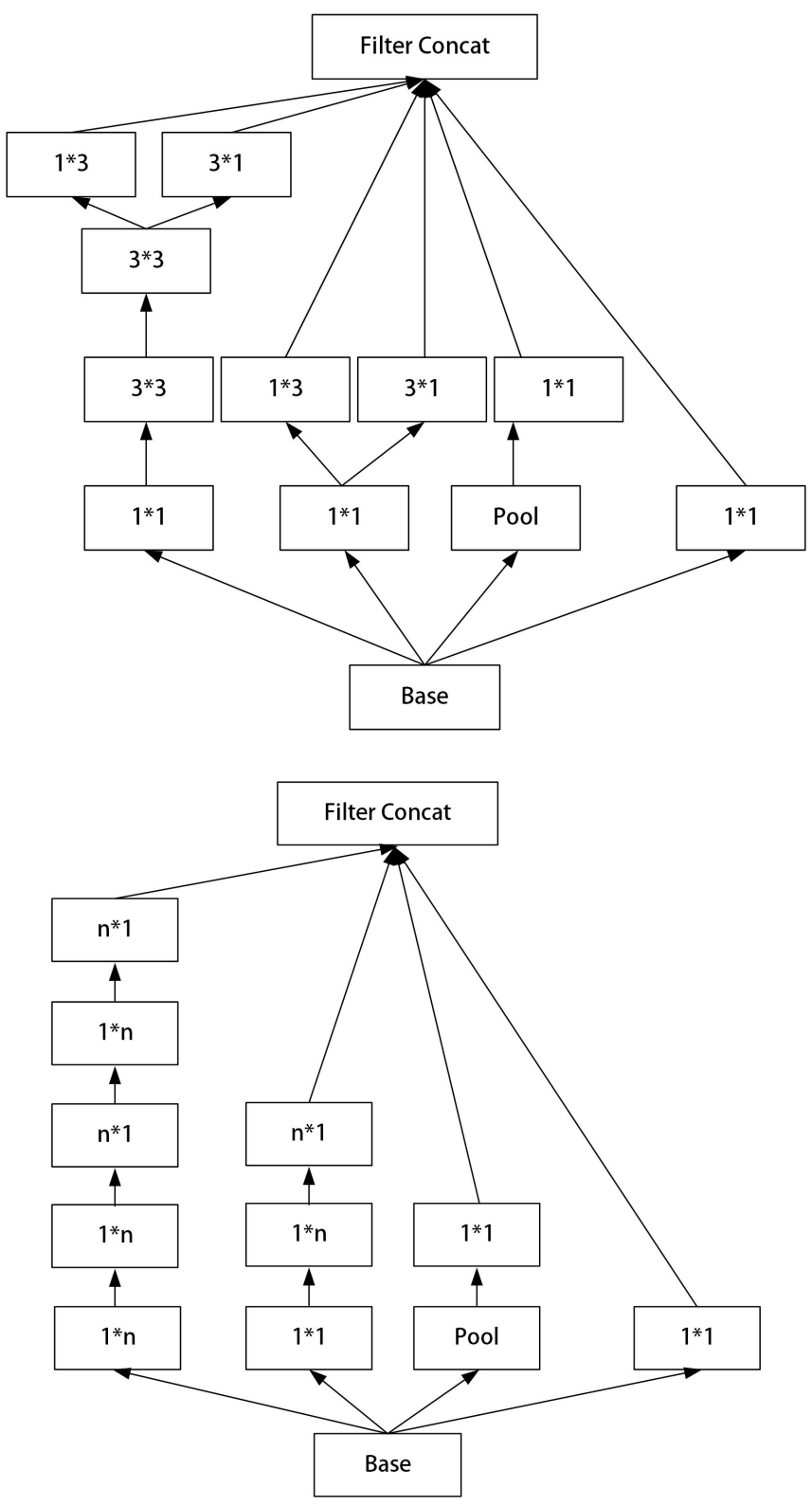

Fig. 3. Inception module in Inception v3.

### *2.2. Model improvements*

In this section, we discuss the enhancements made to the InceptionV3 model to make it particularly suitable for the task of monkeypox disease recognition.

#### *2.2.1. L2 regularization*

In deep learning, we seek models with superior predictive ability and at the same time want to avoid overfitting phenomena. To achieve this goal, regularization techniques are usually employed to control the complexity of the model, thus preventing overfitting and ensuring better generalization ability of the model.

L2 regularization is introduced in our model to mitigate overfitting. Monkeypox disease images can exhibit considerable variation in appearance and we wanted to ensure that our model generalized well to different cases. By applying L2 regularization to the model's weights, we impose constraints that encourage smaller weight values, effectively preventing the model from fitting noise to the training data. This regularization technique plays a crucial role in improving the robustness of the model and its ability to handle variations in monkeypox disease manifestations.

L2 regularization is implemented to limit the complexity of the model by adding the sum of squares of the weight parameters to the loss function. This takes the form of adding one more term to the original loss function: $\frac{1}{2}\lambda\theta_i^2$ , The loss function with the addition of the L2 regular term can then be expressed as:

$$L(\theta) = L'(\theta) + \lambda\sum_{i=1}^{n}\theta_i^2$$

Where $L(\theta)$ is the loss function, $L'(\theta)$ is the original loss, $n$ is the total number of training samples, $\theta_i$ denotes the weight parameter of the model, i.e., each parameter in the neural network and $\lambda$ is the weight of the regularization term, where a larger value will constrain the model complexity to a greater extent and vice versa.

L2 constraints are known for imposing significant penalties on weight vectors that exhibit spiky patterns, favoring more uniform weight parameters. This encourages neural units to consider inputs from upper layers more comprehensively, rather than relying heavily on a select few. When the L2 regularization term is introduced, it tends to reduce the absolute magnitude of weights across the network, ensuring that no excessively large values (such as noise) dominate. In essence, the network leans towards learning smaller and more balanced weights.

It can be seen that the contraction of the weight vector before performing the gradient update at each step is the reason why it is called weight decay. Thus, L2 regularization does allow the weights to become smaller, i.e., it controls the range of values of the parameters to control the capacity of the model, and the lower the capacity of the model the less likely it is to cause overfitting.

#### *2.2.2. Introduction of the SENet module*

In 2017, Ho Jie and his team proposed Squeeze and Excitation (SENet) [3], a new structure for image recognition based on the relationship between feature channels. The central concept here revolves around modeling the inter-channel correlations within feature maps to derive input feature maps and channel-specific weights. Additionally, a novel feature recalibration technique is employed, which dynamically learns the significance of each feature channel. This approach enhances useful features while suppressing less relevant ones for the given task, ultimately contributing to improved model performance and effectiveness. Fig 4 illustrates the network's structural diagram.

The SENet module was incorporated into our model to enhance its feature selection capabilities. Monkeypox skin lesions may have subtle patterns and features that are critical for accurate diagnosis. The SENet module enables our InceptionV3 model to automatically learn the importance of different feature maps or channels in the network. By adaptively recalibrating these feature maps based on their importance, the model can focus on capturing key patterns associated with monkeypox, thus improving its discriminative power. This addition is particularly beneficial in recognizing subtle aspects of monkeypox skin lesions, which can be challenging for traditional CNN architectures.

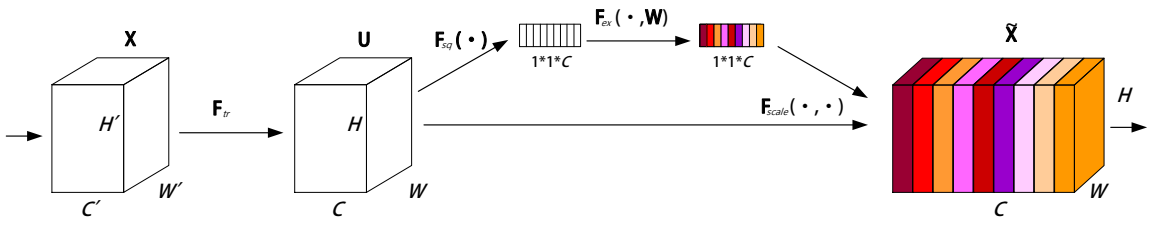

Fig. 4. Network structure diagram of SENet.

Fig 4 illustrates the workflow of the SENet module. The first operation performed is the Squeeze operation, which compresses the features along the spatial dimension. Through global average pooling, this step converts the 2D feature map of each channel into a single scalar value representing the overall activity of that channel. This not only reduces the number of parameters in the model and reduces the computational complexity, but also allows the model to capture global information between channels, laying the foundation for subsequent operations.

Next is the Excitation operation, whose mechanism is similar to the gating mechanism in recurrent neural networks. It is based on the global features provided by the Squeeze operation and learns to generate a set of weights for each channel by means of a simple neural network (typically containing two fully connected layers and a Sigmoid activation function). This process allows the model to dynamically adjust the weights for each channel, which in turn reinforces or suppresses specific features. This automatic reinforcement and suppression of features allows the network to be more focused and effective in dealing with specific tasks.

The final Scale operation serves to recalibrate each channel in the original feature map using the weights derived from the Excitation operation, completing the adjustment of features in the channel dimension. The combined effect of this series of operations not only improves the performance of the model but also enhances its discriminative ability in specific tasks, such as monkeypox lesion recognition.

Now, the above SENet mechanism is embedded into the InceptionV3 model after channel compression, and the realization process is shown in Fig 5. Where  is the input image size, C is the number of channels, Global pooling is the global average pooling, FC is the fully connected layer, and Sigmoid is the activation function.

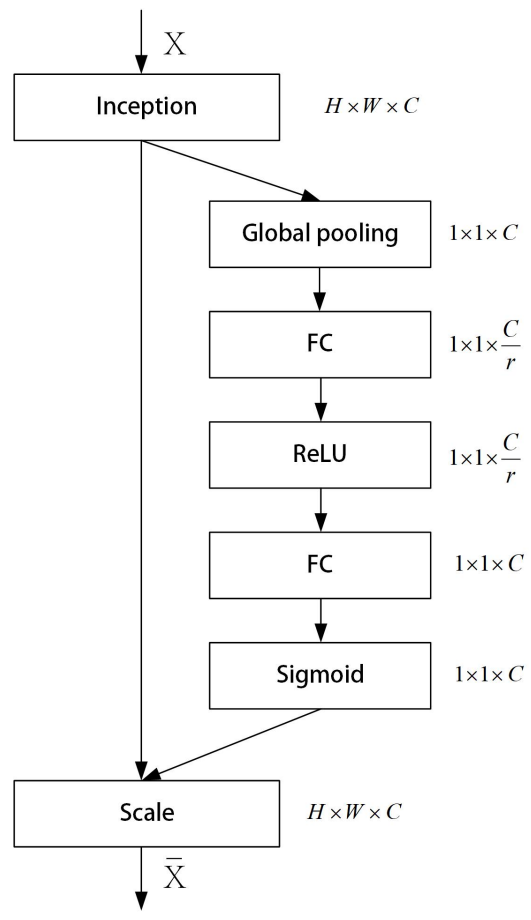

Fig. 5. Inception model after embedding SENet module.

By incorporating L2 regularization and SENet modules into our InceptionV3 model, we have customized it to excel in the challenging task of monkeypox disease recognition. Together, these enhancements improve the model's ability to handle diverse cases and capture key disease-related patterns, ultimately leading to more accurate and robust diagnoses.

## 3. Model complexity analysis

This study focuses on evaluating the complexity of the enhanced SE-InceptionV3+L2 model compared to a state-of-the-art (SOTA) model[5] and several other mainstream deep learning models. We analyzed the architecture and complexity of these models in detail using the PyTorch framework and the torchsummary tool. All models are evaluated with the same input size to ensure consistency and comparability of results.

The complexity assessment focuses on the total number of parameters of the model, the number of trainable and non-trainable parameters, and the estimated total size of the model. This assessment aims to understand the computational resource and storage requirements of different models, especially the potential for application in resource-constrained environments. For example, VGG-16 is more suitable for applications with high accuracy requirements due to its high number of parameters and memory requirements, while medium-sized models such as SE-InceptionV3+L2 strike a good balance between performance and computational efficiency.

Table 1 presents the results of the complexity analysis of the compared models, highlighting their significant differences in complexity. These data emphasize the importance of choosing the right model based on the accuracy, speed, and resource constraints required for a particular task.

Table 1

Analysis of Model Complexity

| Network Models | Total Params | Input Size (MB) | Forward/Backward Pass Size (MB) | Parameter size (MB) | Estimated total size (MB) |
|---|---|---|---|---|---|
| SE-InceptionV3+L2 | 24,381,299 | 1.02 | 329.34 | 93.01 | 423.37 |
| ResNet50V2 | 25,549,416 | 1.02 | 444.19 | 97.46 | 542.67 |
| VGG-16 | 134,276,932 | 0.57 | 218.52 | 512.23 | 731.32 |
| InceptionV1 | 6,660,244 | 0.57 | 71.97 | 25.41 | 97.95 |
| SE-DenseNet | 6,957,572 | 0.57 | 4854.11 | 26.54 | 4881.22 |
| ResNeXt-50 | 37,574,724 | 0.57 | 379.37 | 143.34 | 523.28 |
| ResNet+DenseNet | 95,008,356 | 1.02 | 1204.51 | 362.43 | 1567.96 |
| YOLOv5C3 | 25,415,992 | 1.02 | 237.59 | 96.95 | 335.57 |
| InceptionV3 | 23,834,568 | 1.02 | 292.54 | 90.92 | 384.49 |
| DenseNet121-TL | 6,962,058 | 0.57 | 383.87 | 26.56 | 411.01 |

The above analysis shows that despite the complex structure of the SE-InceptionV3+L2 model, we pay special attention to the optimization of computational efficiency and complexity in the design. This balance makes the SE-InceptionV3+L2 model not only perform well in resource-limited environments, but also lays the foundation for subsequent performance evaluation.

## 4. Experiment

In this section, the following parts are as follows: first, a brief overview and introduction of the dataset are given; second, the configuration and parameter settings of the experiments are described in detail; then, the ablation experiments are conducted and analyzed; and finally, the validity of the proposed method is fully verified by comparison experiments with other methods.

### *4.1. Experimental dataset*

The experimental dataset is the Kaggle monkeypox disease dataset [4,5], co-created in 2022 by NAFISA6615 and three collaborators. This dataset was compiled by collecting and processing images from various web crawls, such as news portals, websites, and publicly accessible case reports. All images were validated or reviewed by specialized dermatologists, ensuring their medical relevance and accuracy. The original dataset consisted of a total of 228 images, divided into two categories: 102 images of monkeypox disease and 126 images of similar non-monkeypox cases (e.g., chickenpox and measles). This distribution reflects the challenge in differentiating monkeypox from other visually similar skin conditions, emphasizing the necessity for a highly accurate classification model. The proportion of images in the monkeypox and other categories was 44.7% and 55.3%, respectively, showcasing the dataset's balanced nature. Each image in the dataset has a resolution of 224×224 pixels.

In order to enhance the diversity of the dataset, we applied a variety of image data enhancement methods including rotation, panning, reflection, cropping, hue adjustment, saturation, contrast, luminance dithering, noise addition, and scaling. These enhancement techniques were chosen to simulate the diverse variations of skin lesions that may occur in the real world and to improve the robustness of the model in dealing with different skin lesions. For example, through rotation and panning operations, the model can learn to recognize lesions in different positions and orientations; while adjusting the hue and brightness of the image helps the model to adapt to different lighting conditions and skin tones. Using these data enhancement techniques, we obtained an extended dataset containing 1,428 images in the monkeypox disease category and 1,764 images in other categories. Such data enhancement strategies not only enriched the dataset but also provided more comprehensive training for the model, thus enhancing its recognition ability and adaptability.

This dataset presents unique challenges for disease recognition models due to its diverse representation of skin lesions and the subtle differences between monkeypox and other similar diseases. The variability in appearance, size, and color of the lesions within the monkeypox category itself poses a significant challenge. Moreover, the inclusion of other diseases with similar symptoms requires the model to not only identify monkeypox-specific features but also to distinguish them from other diseases with high accuracy. To address these challenges, our model is designed to be sensitive to the subtle variations in skin lesions while being robust enough to differentiate between monkeypox and other conditions effectively. This sensitivity and specificity are crucial, given the high stakes of medical diagnosis where misclassification can have serious implications. For the experiment, all image data were divided into training and test sets according to a 4:1 ratio. Fig 6 presents three original images from each category of the monkeypox disease and other categories.

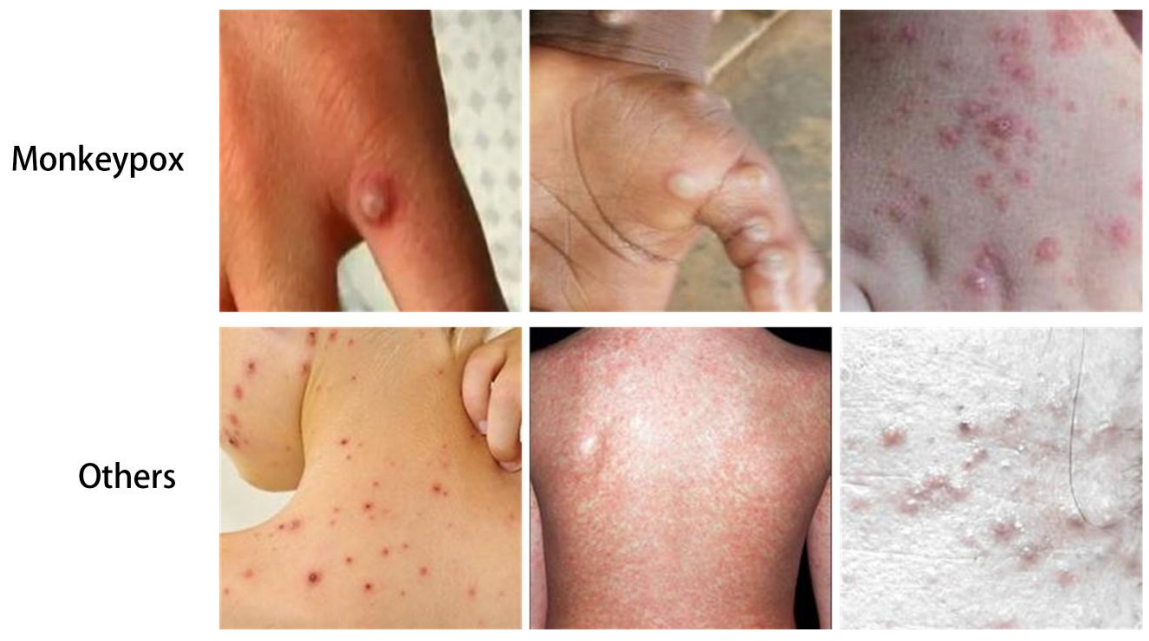

Fig. 6. Example of experimental dataset.

In addition, we recognize the potential biases and limitations that exist in the dataset. As the data were collected from publicly available sources primarily through web crawlers, they may not be fully representative of all populations, particularly with respect to ethnicity and skin type diversity. Also, the relatively small size of the dataset may not be sufficient to cover all manifestations of monkeypox in different populations, which may affect the generalization ability of the model. To address these challenges, we have improved the diversity of the dataset through data enhancement techniques, while in the future we plan to collect more diverse and extensive case data to improve the accuracy and representativeness of the model.

### *4.2. Experimental design and sensitivity analysis*

#### *4.2.1. Experimental design and model convergence monitoring*

The experimental operating system was Windows 10 64-bit operating system and the computer was equipped with a GPU of RTX3050Ti with CUDA for training. The programming language was Python and the deep learning framework was PyTorch. The learning rate was set to 0.0001, batch size is 16, and 50 epochs of training were performed. The training was performed using Adam Optimizer. To ensure the consistency and reproducibility of the experimental results, fixed random seeds 42 were set in all experiments.

To ensure that the improved SE-InceptionV3 model converged effectively, we monitored the two main metrics, cross-entropy loss function and accuracy. One of the signs of model convergence is a continuous decrease in the loss value of the validation set. In addition, an early stopping strategy was implemented, i.e., we stopped training if the validation set loss did not show significant improvement within 10 consecutive training epochs. This measure helps prevent overfitting and ensures the generalization ability of the model.

#### *4.2.2. Sensitivity analysis of hyperparameters*

In order to deeply explore the sensitivity of the SE-InceptionV3+L2 model to hyperparameters, this study conducts a careful and systematic analysis of three key hyperparameters, namely, learning rate, batch size, and training epoch. Hyperparameters play a crucial role in model training as they directly affect the learning process and performance stability of the model. In the following experiments, only one hyperparameter is adjusted at a time while keeping the other hyperparameters fixed to observe the impact of each hyperparameter individually.

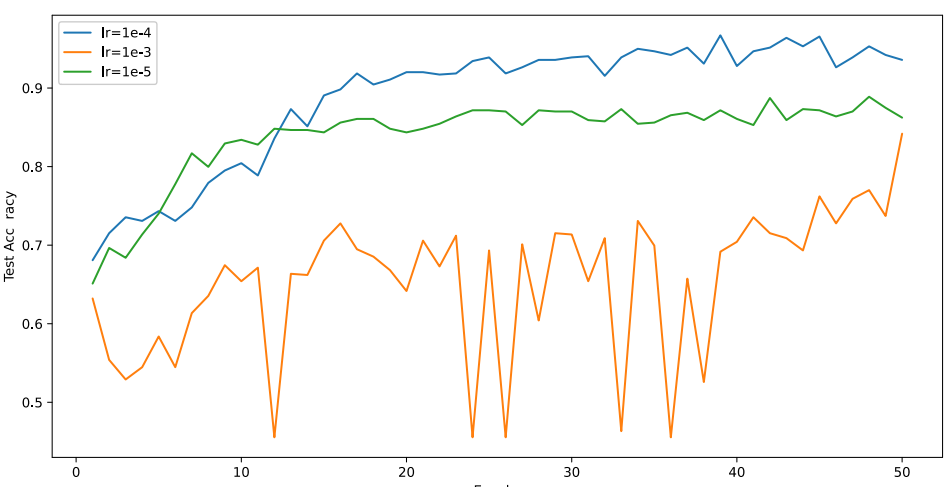

Fig. 7. Accuracy under different learning rates.

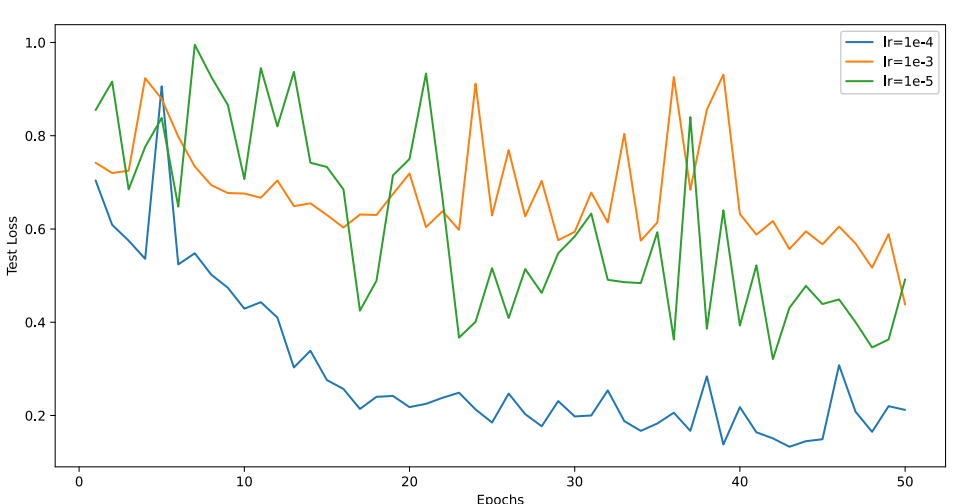

Fig. 8. Loss under different learning rates.

First, we tested the learning rate. Learning rate is an important parameter that determines the update step of the model, which directly affects the convergence speed and stability of the model. We tried three different learning rates, 1e-3, 1e-4, and 1e-5. At the high learning rate of 1e-3, the model converges too fast, which leads to a higher validation loss and accuracy fluctuation as it is more difficult to reach the global optimal solution during the learning process. In contrast, the learning rate of 1e-4 provides a more balanced solution, achieving a balance of fast convergence and higher accuracy. In contrast, at the low learning rate of 1e-5, the model converges too slowly and may require more training epochs to achieve similar performance.

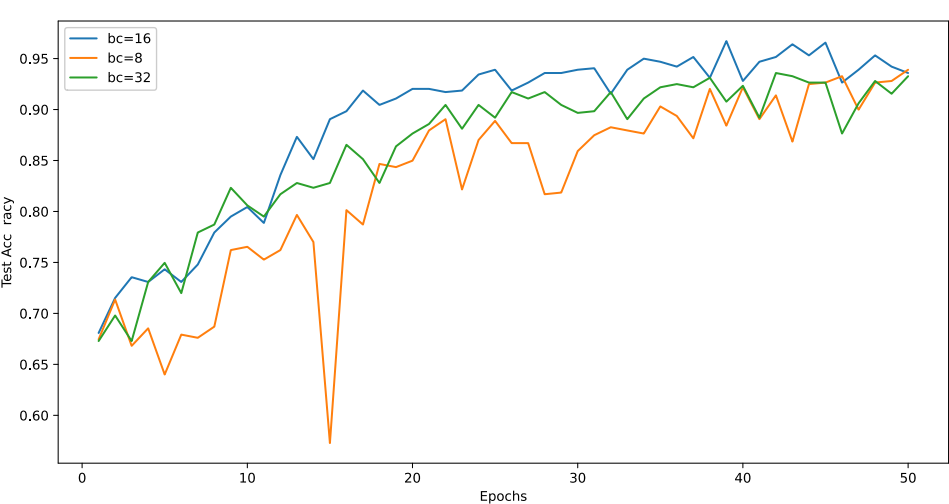

Fig. 9. Accuracy at different batch sizes.

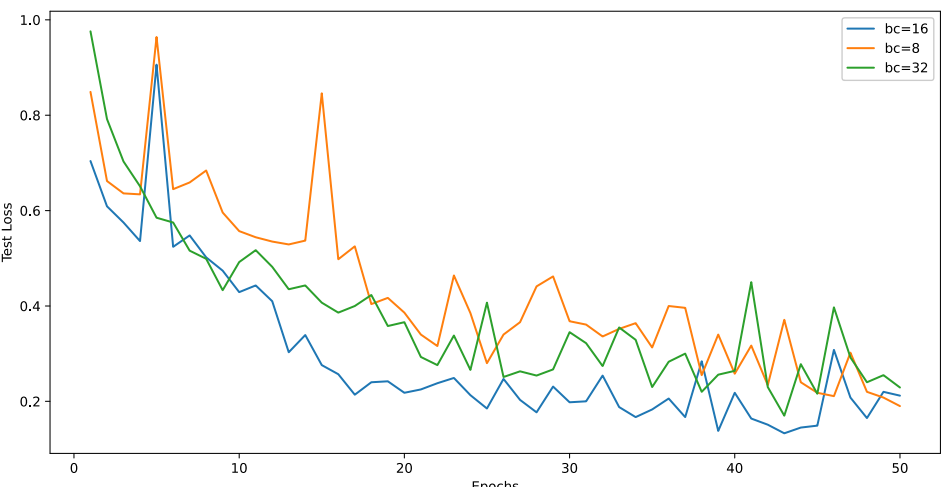

Fig. 10. Loss at different batch sizes.

Next, we examined the effect of batch size on model performance. Batch size determines the number of samples processed by the network during each training session and has a direct impact on memory usage and gradient estimation accuracy. We tested three different batch sizes, 8, 16, and 32, respectively. The experimental results show that with a batch size of 8, the gradient update of the model is noisier, which helps the model avoid falling into local minima but also makes the learning process more unstable. On the contrary, with a batch size of 32, although the training progress is smoother, the updating is less noisy, resulting in slower model learning. In contrast, with a batch size of 16, the model exhibits the best performance, effectively balancing the stability of gradient estimation with computational efficiency, and the loss curve shows a more stable training process and potentially better generalization ability.

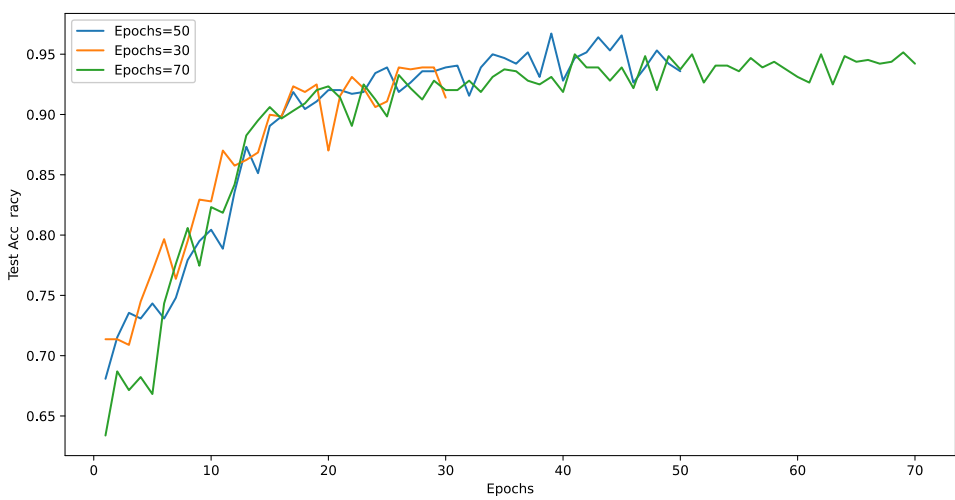

Fig. 11. Accuracy under different Epochs.

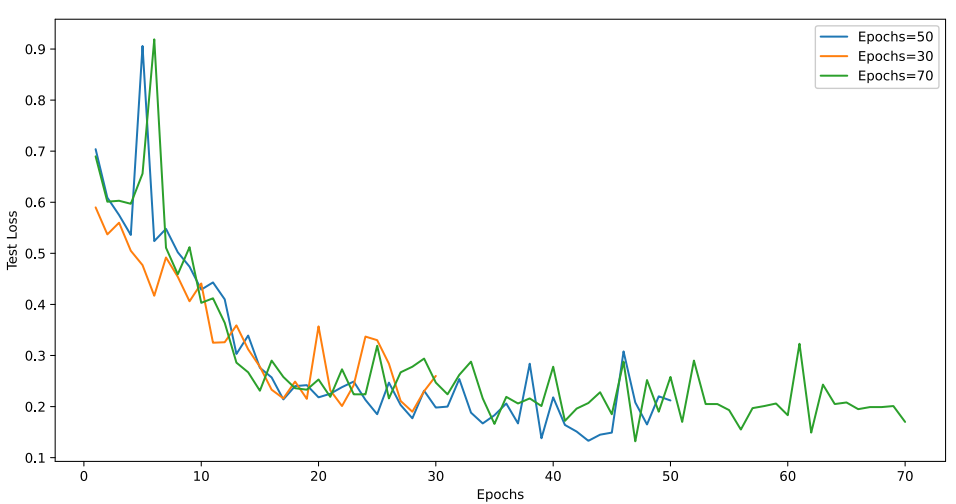

Fig. 12. Loss under different Epochs.

Finally, we analyzed the effect of the training epoch on the model performance. The training epoch defines the number of times the model passes through the entire training dataset in its entirety and has a significant impact on model convergence and overfitting. We tested the model for 30, 50, and 70 training epochs. 30 epochs showed signs of underfitting, indicating that the training time was not sufficient for the model to learn adequately. In contrast, at 50 training epochs, the model's performance peaked, showing the highest precision and recall. Increasing to 70 epochs did not observe a significant performance improvement, suggesting that continuing to increase the number of training cycles may lead to overfitting.

Overall, the results of these experiments emphasize the importance of proper tuning of hyperparameters to achieve optimal model performance. Through in-depth analyses of the learning rate, batch size, and training epochs, we can better understand how they affect the learning and generalization process of deep learning models. These findings not only guide further optimization of the SE-InceptionV3+L2 model but also provide valuable references for future research, especially in exploring the specific impact of hyperparameter tuning and model architecture changes on target detection performance.

### *4.3. Ablation experimental*

In order to demonstrate the validity of the monkeypox disease identification model proposed in this paper and at the same time to analyse its performance in the convergence process, we conducted several experiments, including a comparison between our proposed model and the InceptionV3 [2] model. In the InceptionV3 model, we added the SENet module and L2 regularization, respectively, and applied them to the test set. The accuracy of the four models based on InceptionV3 is shown in Table 1. Fig 7 shows the variation curves of the accuracy of each model and Fig 8 shows the variation curves of Loss, Precision, Recall, and F1.

Finally, we obtained the accuracy rates of 96.71%, 94.21%, 95.46%, and 94.37% for four different configurations of models in recognizing monkeypox disease, respectively. The experimental results show that when we added the SE attention mechanism module and L2 regularization, the recognition accuracy improved by 1.25% and 0.16%, respectively, compared to the original InceptionV3 model. This result clearly shows that these improvements enhance the model's recognition accuracy for monkeypox disease to some extent.

Table 2

Ablation experimental results based on InceptionV3

| Network Models | Test_acc(%) | Test_loss | Test_Precision | Test_Recall | Test_F1 |
| --- | --- | --- | --- | --- | --- |
| SE-InceptionV3+L2 | 96.71 | 0.138 | 0.954 | 0.95 | 0.952 |
| InceptionV3+L2 | 94.37 | 0.187 | 0.935 | 0.935 | 0.935 |
| SE-InceptionV3 | 95.46 | 0.130 | 0.924 | 0.92 | 0.922 |
| InceptionV3 | 94.21 | 0.238 | 0.933 | 0.934 | 0.933 |

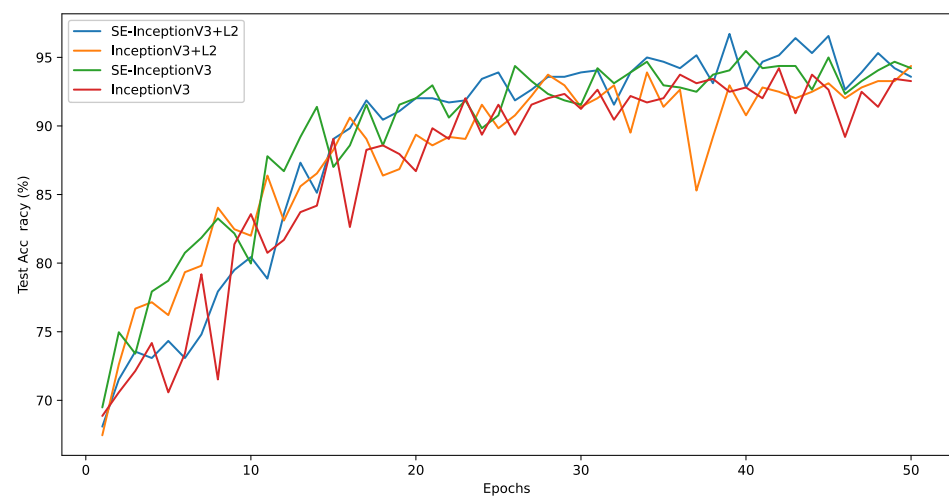

Fig. 13. Accuracy curve of each model.

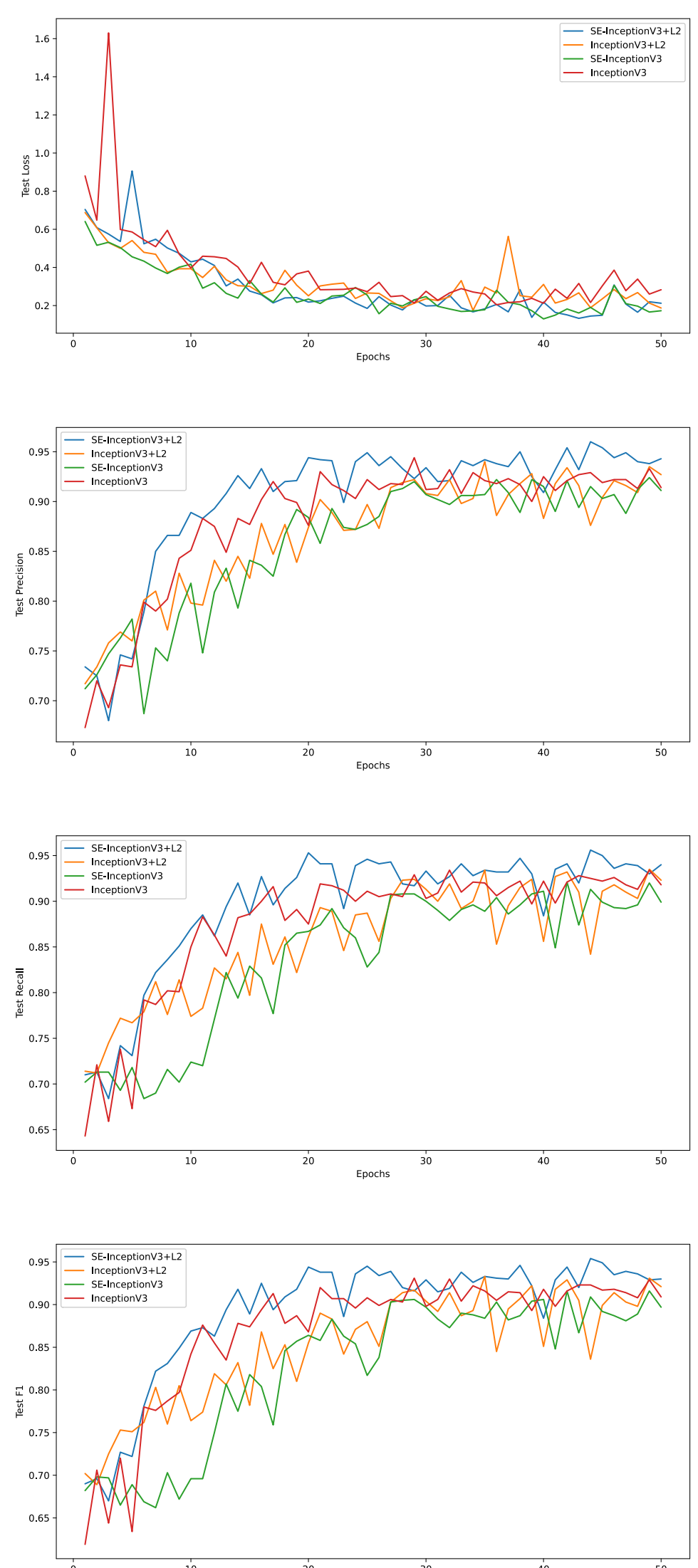

Fig. 14. Loss, Precision, Recall, and F1 curves for each model.

According to the display in Fig 13 and Fig 14, the proposed SE-InceptionV3+L2 model performs better in Accuracy relative to the other models after 30 epochs of training and shows a trend of gradual convergence, indicating that the model has made significant progress in the classification task. In addition, the model's Precision, Recall, and F1 also outperform the other models throughout the training process. It means that the model has excellent classification performance in both positive and negative categories, as well as being able to find the positive categories efficiently and reduce misclassification. Also, the model exhibits lower loss values and remains relatively stable after 20 epochs of training. Low loss values indicate that the model is able to fit better on the training data while maintaining stability indicates that the model is less prone to overfitting. These results clearly demonstrate the performance advantages of our proposed model during training.

We argue that the SENet module improves the model's ability to represent features through its unique channel attention mechanism. By giving the model the ability to emphasize the most relevant features to the task and suppress less important features, the SENet module makes the model more accurate in dealing with monkeypox disease features. On the other hand, L2 regularization reduces the likelihood of overfitting by limiting the size of the model weights, making the model perform more robustly on unknown data.

Combining these observations, we can see that the improved model performs better in terms of Accuracy, Loss, Precision, Recall, and F1 score. Although there are some fluctuations in accuracy in the later stages of training, overall the model demonstrates better results, which confirms the positive impact of the SENet module and L2 regularization on improving model performance.

### *4.4. Comparison experiment*

In the comparative analysis conducted to verify the recognition performance of the improved model, the enhanced SE-InceptionV3+L2 model is ultimately compared and analyzed against ResNet50V2 [16], VGG-16 [19], InceptionV1 [18], SE-DenseNet [8], ResNeXt-50 [17], ResNet+DenseNet [7], YOLOv5C3 [9], and the state-of-the-art DenseNet121-TL[5] model. Table 2 presents the comparison results of accuracy and loss for each model, while Fig 15 and Fig 16 illustrate the curve comparisons.

Table 3

Comparison of modeling results

| Network Models | Test_acc(%) | Test_loss | Test_Precision | Test_Recall | Test_F1 |
|---|---|---|---|---|---|
| SE-InceptionV3+L2 | 96.71 | 0.138 | 0.954 | 0.95 | 0.952 |
| ResNet50V2 | 88.26 | 0.440 | 0.894 | 0.895 | 0.894 |
| VGG-16 | 91.86 | 0.318 | 0.883 | 0.867 | 0.875 |

| InceptionV1 | 94.52 | 0.798 | 0.915 | 0.899 | 0.907 |
|---|---|---|---|---|---|
| SE-DenseNet | 82.63 | 1.117 | 0.802 | 0.82 | 0.811 |
| ResNeXt-50 | 84.35 | 0.901 | 0.877 | 0.859 | 0.868 |
| ResNet+DenseNet | 86.23 | 0.504 | 0.86 | 0.858 | 0.859 |
| YOLOv5C3 | 91.86 | 0.252 | 0.906 | 0.9 | 0.903 |
| DenseNet121-TL | 95.09 | 0.162 | 0.954 | 0.959 | 0.949 |

We chose the highest accuracy in 50 epoch training and its corresponding loss, and the comparison results are shown in Table 2, from which it can be seen that the improved SE-InceptionV3+L2 model is able to achieve an accuracy of 96.71% and the loss is significantly lower than that of other models.

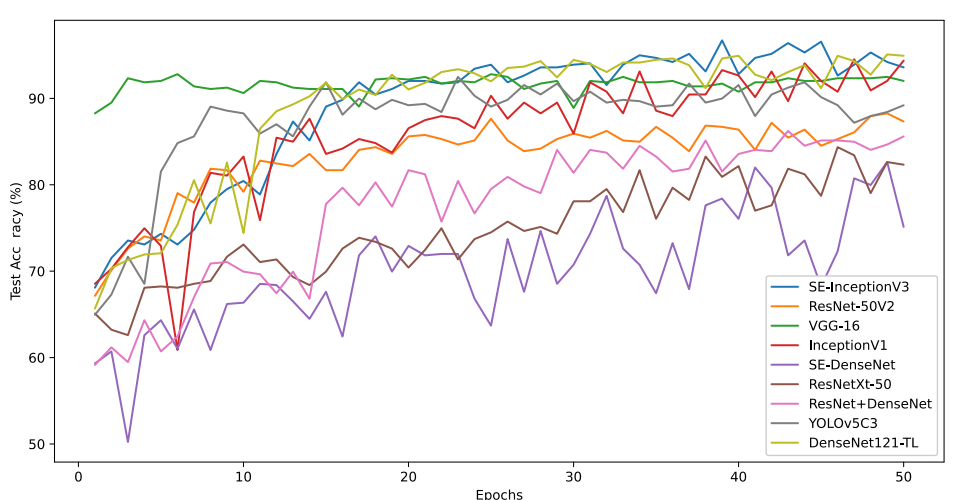

Fig. 15. Comparison of the Accuracy of validation sets of each classification network model.

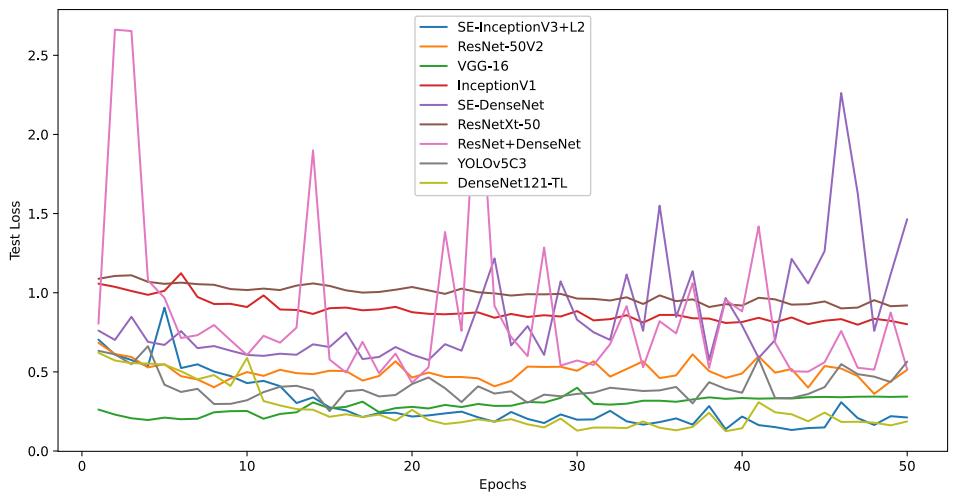

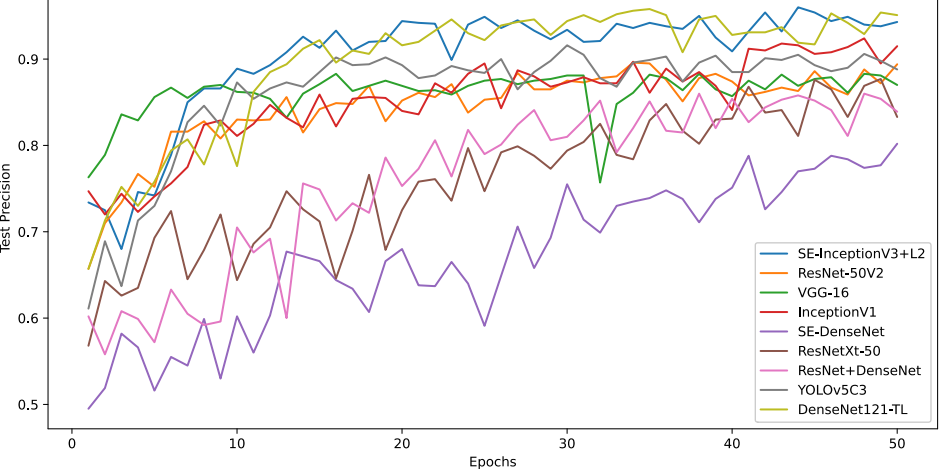

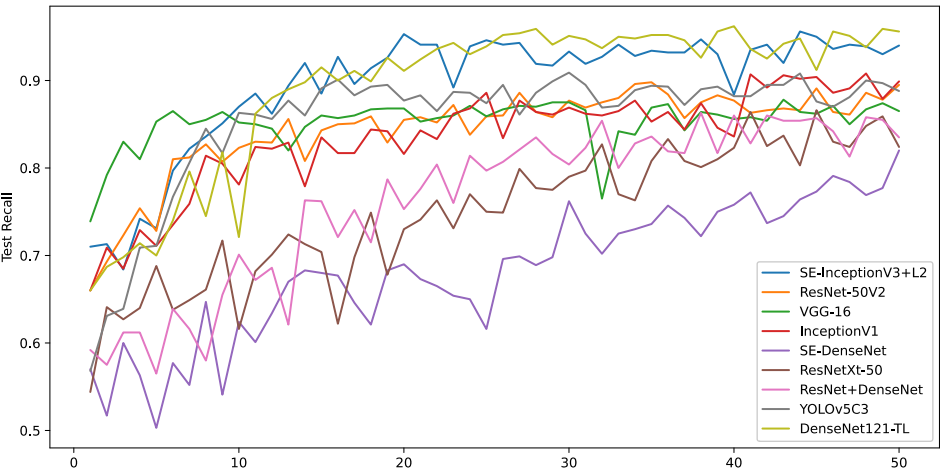

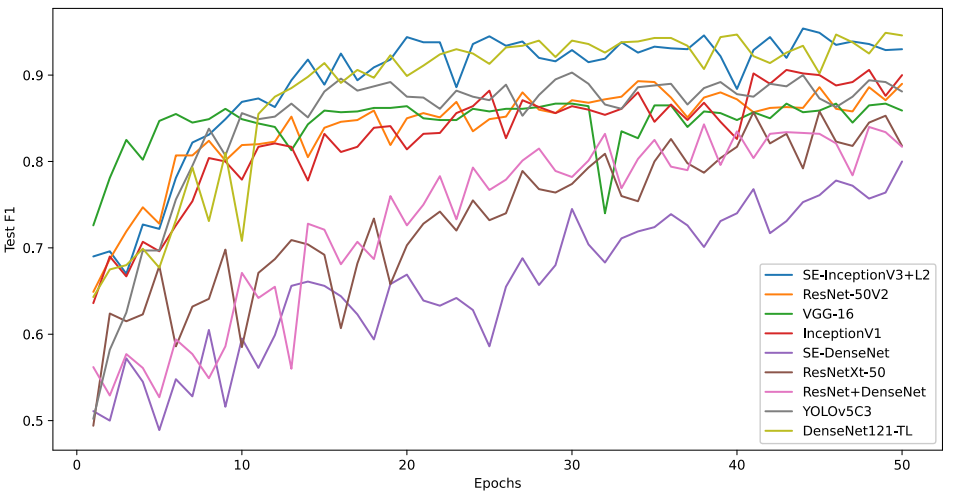

Fig. 16. Comparison of the validation set Loss, Precision, Recall, and F1 for each classification network model.

In the above graph, the horizontal coordinate indicates the number of iterations, and the vertical coordinate indicates the changes in metrics such as accuracy. As mentioned earlier, the number of training iterations of the model is 50. observing the graphs, it can be seen that when conducting classification experiments, when the number of iterations reaches about 10 times, the performance of the model's Precision, Recall, and F1 starts to outperform other models and reaches a relatively stable interval. Meanwhile, when the number of iterations reaches about 20 times, the Accuracy and Loss rate of our proposed model maintains a stable state.

Combining the results in Table 2, Fig 15, and Fig 16, the improved SE-InceptionV3 model exhibits an accuracy of 96.71% and is significantly better than the other models in terms of Precision, Recall, and F1 metrics. In addition, its loss value is significantly lower than the other models. Compared to the SOTA model DenseNet121-TL, Accuracy, Loss, and Precision performed similarly in the first thirty-five epochs; however, after thirty-five epochs, the model slightly outperformed the SOTA model. In terms of Recall and F1 metrics, the proposed model performs similarly to the SOTA model. The experimental results show that the SE-InceptionV3 model proposed in this paper has a significant advantage in improving the recognition rate compared to other algorithms.

To assess the SE-InceptionV3+L2 model's effectiveness, we engaged in a thorough performance evaluation. A principal indicator was the model's capacity to discern between monkeypox and other similar conditions with precision. The confusion matrix depicted in Fig. 17 represents the outcomes from the final epoch of training, showcasing the model's adeptness in classification.In Fig. 18, we provide a series of sample images from the training process that show how the model performs during the classification process. These examples show the model correctly classifying and incorrectly classifying situations, presented as images.

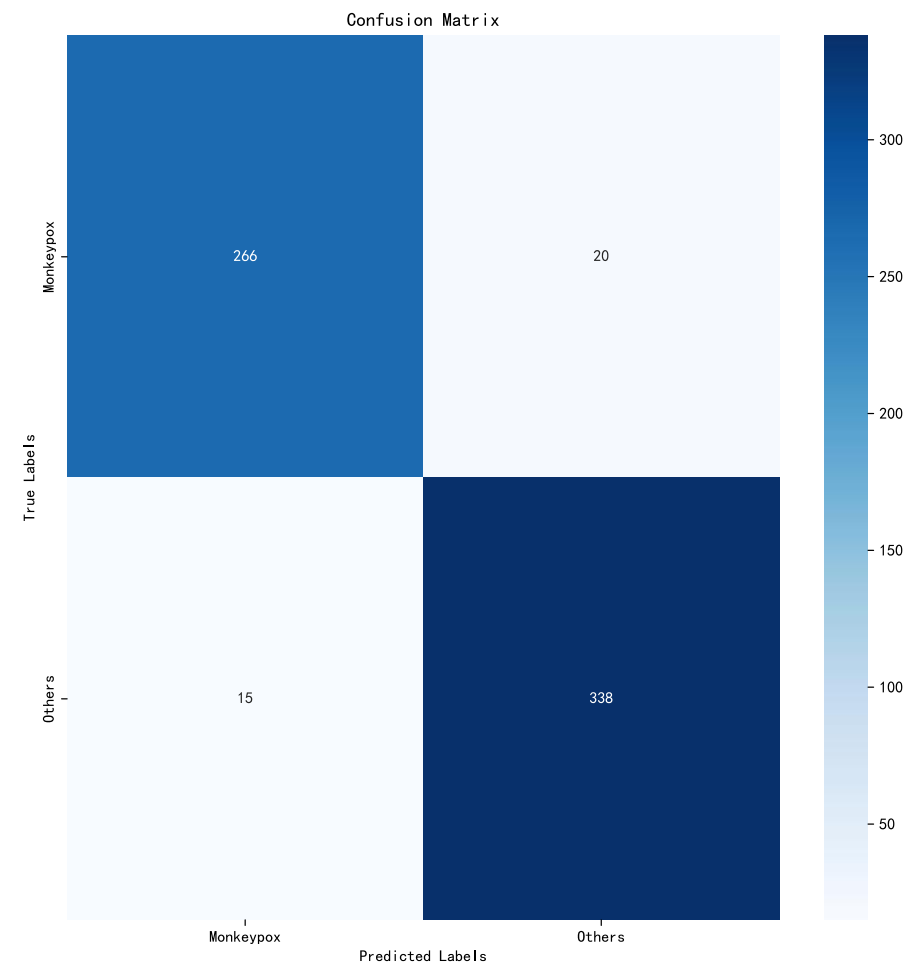

Fig. 17. The Confusion Matrix of SE Conception V3+L2 Model.

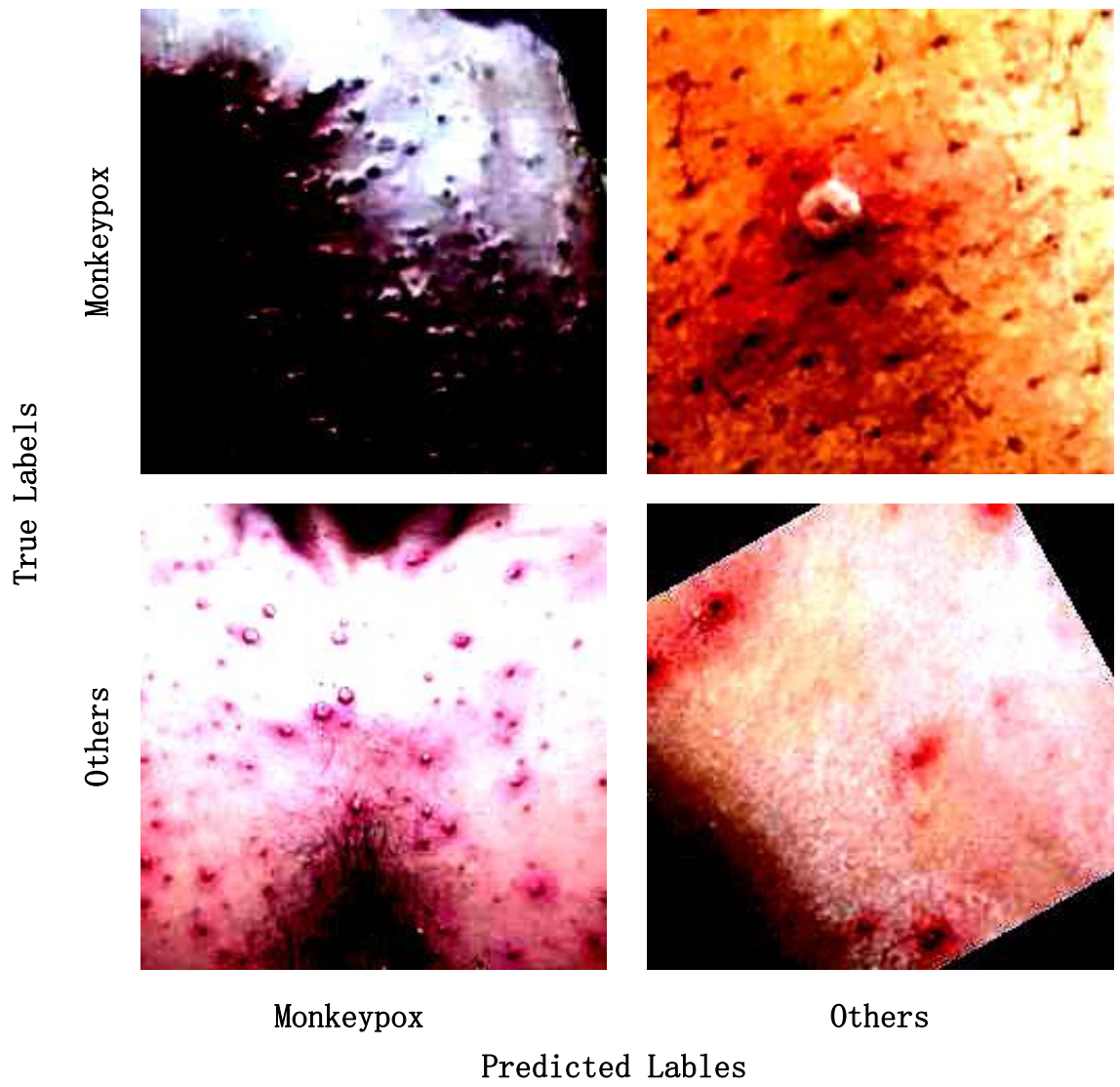

Fig. 18. Visualize incorrect and correct classification during training.

The confusion matrix analysis for the SE-InceptionV3+L2 model indicated that it correctly identified 266 cases of monkeypox (True Positives) and 338 non-monkeypox cases (True Negatives), demonstrating impressive performance in distinguishing between the two conditions. The model misclassified 15 cases of monkeypox as non-monkeypox (False Negatives) and misjudged 20 non-monkeypox cases as monkeypox (False Positives). Consequently, the model achieved a precision of 93%, a recall rate of 94.7%, and an F1 score of 93.8%. These metrics reflect the model's balanced accuracy in detecting monkeypox and non-monkeypox cases. Notably, the low rate of False Negatives underscores the model's potential in reducing undetected monkeypox cases, which is crucial for public health management. These results demonstrate the superior diagnostic accuracy of our model using convolutional neural networks for medical image analysis compared to traditional methods.

It is also worth noting that although our SE-InceptionV3+L2 model performs well on several metrics, there is still some uncertainty in the model prediction. This uncertainty may stem from the diversity of datasets, the limitation of training samples, and the inherent complexity of deep learning models. In order to quantify this uncertainty, we performed several repetitive experiments on the model and recorded the performance fluctuations in each experiment. We found that despite the overall accuracy of the model, metrics such as accuracy, loss, precision, recall, and F1 score show some degree of variation across different experimental repetitions. This suggests that the interpretation and use of model predictions in practical applications should take these potential uncertainties into account.

Reporting model uncertainty is an important part of our research, and for future work we plan to explore further ways to reduce this uncertainty, such as using larger or more representative datasets, improving the model structure, or applying more advanced training techniques. These measures are expected to improve the stability and reliability of the models under different conditions.

### *4.5. Statistical significance analysis*

Ultimately, this study provides an in-depth performance evaluation of a wide range of state-of-the-art target detection models taking into account key performance metrics such as Accuracy, Loss, Recall, Precision, and F1 score. The models we analysed involve DenseNet + SE-Net, Inception v1, Inception v3, Inception v3 + L2, ResNeXt-50, ResNet + DenseNet, ResNet50v2, SE-InceptionV3, SE- InceptionV3+L2, VGG-16, and Yolov5C3. All models were evaluated on their respective 50-sample-capacity datasets, totaling 11 data sets. We applied the statistical technique of analysis of variance (ANOVA) to explore the statistical significance of the performance differences between these models, and the results are summarised and presented in the table below.

Table 4

ANOVA Results for Accuracy

| **Analysis item** | **item** | **sample capacity** | **AVG** | **SD** | ***F*** | ***p*** |
|---|---|---|---|---|---|---|
| Test Accuracy | DenseNet+SE-Net | 50 | 0.7 | 0.07 | 52.645 | 0.000** |
| | Inception v1 | 50 | 0.86 | 0.07 | | |
| | Inception v3 | 50 | 0.89 | 0.06 | | |
| | InceptionV3+L2 | 50 | 0.88 | 0.06 | | |
| | ResNeXt-50 | 50 | 0.75 | 0.06 | | |
| | ResNet+DenseNet | 50 | 0.77 | 0.08 | | |
| | ResNet50v2 | 50 | 0.83 | 0.05 | | |
| | SE-InceptionV3 | 50 | 0.87 | 0.08 | | |
| | SE-InceptionV3+L2 | 50 | 0.89 | 0.07 | | |
| | VGG-16 | 50 | 0.87 | 0.01 | | |
| | Yolov5C3 | 50 | 0.87 | 0.06 | | |
| | Densenet121+TL | 50 | 0.89 | 0.08 | | |
| | in total | 600 | 0.84 | 0.09 | | |

* p<0.05 ** p<0.01

Table 5

ANOVA Results for Loss

| **Analysis item** | **item** | **sample capacity** | **AVG** | **SD** | ***F*** | ***p*** |
|---|---|---|---|---|---|---|
| Test Loss | DenseNet+SE-Net | 50 | 0.85 | 0.33 | 100.173 | 0.000** |
| | Inception v1 | 50 | 0.88 | 0.07 | | |
| | Inception v3 | 50 | 0.27 | 0.12 | | |
| | InceptionV3+L2 | 50 | 0.32 | 0.12 | | |
| | ResNeXt-50 | 50 | 0.99 | 0.05 | | |
| | ResNet+DenseNet | 50 | 0.87 | 0.52 | | |
| | ResNet50v2 | 50 | 0.5 | 0.06 | | |
| | SE-InceptionV3 | 50 | 0.3 | 0.17 | | |
| | SE-InceptionV3+L2 | 50 | 0.32 | 0.12 | | |
| | VGG-16 | 50 | 0.29 | 0.05 | | |
| | Yolov5C3 | 50 | 0.41 | 0.09 | | |
| | Densenet121+TL | 50 | 0.26 | 0.14 | | |
| | in total | 600 | 0.52 | 0.34 | | |

* p<0.05 ** p<0.01

Table 6

ANOVA Results for Recall

| **Analysis item** | **item** | **sample capac** | **AVG** | **SD** | ***F*** | ***p*** |
|---|---|---|---|---|---|---|

| | | ity | | | | |
|---|---|---|---|---|---|---|
| Test Recall | DenseNet+SE-Net | 50 | 0.68 | 0.08 | 39.874 | 0.000** |
| | Inception v1 | 50 | 0.84 | 0.06 | | |
| | Inception v3 | 50 | 0.87 | 0.07 | | |
| | InceptionV3+L2 | 50 | 0.86 | 0.06 | | |
| | ResNeXt-50 | 50 | 0.75 | 0.08 | | |
| | ResNet+DenseNet | 50 | 0.76 | 0.09 | | |
| | ResNet50v2 | 50 | 0.84 | 0.05 | | |
| | SE-InceptionV3 | 50 | 0.84 | 0.08 | | |
| | SE-InceptionV3+L2 | 50 | 0.9 | 0.07 | | |
| | VGG-16 | 50 | 0.85 | 0.03 | | |
| | Yolov5C3 | 50 | 0.85 | 0.07 | | |
| | Densenet121+TL | 50 | 0.88 | 0.08 | | |
| | in total | 600 | 0.83 | 0.09 | | |

* p<0.05 ** p<0.01

Table 7

ANOVA Results for Precision

| **Analysis item** | **item** | **sample capacity** | **AVG** | **SD** | ***F*** | ***p*** |
|---|---|---|---|---|---|---|
| Test Precision | DenseNet+SE-Net | 50 | 0.67 | 0.08 | 51.085 | 0.000** |
| | Inception v1 | 50 | 0.85 | 0.05 | | |
| | Inception v3 | 50 | 0.88 | 0.07 | | |
| | InceptionV3+L2 | 50 | 0.87 | 0.06 | | |
| | ResNeXt-50 | 50 | 0.76 | 0.08 | | |
| | ResNet+DenseNet | 50 | 0.76 | 0.09 | | |
| | ResNet50v2 | 50 | 0.85 | 0.05 | | |
| | SE-InceptionV3 | 50 | 0.86 | 0.07 | | |
| | SE-InceptionV3+L2 | 50 | 0.9 | 0.07 | | |
| | VGG-16 | 50 | 0.86 | 0.03 | | |
| | Yolov5C3 | 50 | 0.86 | 0.07 | | |
| | Densenet121+TL | 50 | 0.89 | 0.07 | | |
| | in total | 600 | 0.83 | 0.09 | | |

* p<0.05 ** p<0.01

Table 8

ANOVA Results for F1

| **Analysis item** | **item** | **sample capacity** | **AVG** | **SD** | ***F*** | ***p*** |
|---|---|---|---|---|---|---|
| Test F1 Score | DenseNet+SE-Net | 50 | 0.66 | 0.08 | 41.853 | 0.000** |
| | Inception v1 | 50 | 0.83 | 0.06 | | |
| | Inception v3 | 50 | 0.87 | 0.08 | | |
| | InceptionV3+L2 | 50 | 0.85 | 0.07 | | |
| | ResNeXt-50 | 50 | 0.74 | 0.08 | | |
| | ResNet+DenseNet | 50 | 0.73 | 0.1 | | |
| | ResNet50v2 | 50 | 0.84 | 0.05 | | |
| | SE-InceptionV3 | 50 | 0.83 | 0.09 | | |
| | SE-InceptionV3+L2 | 50 | 0.89 | 0.07 | | |
| | VGG-16 | 50 | 0.85 | 0.03 | | |
| | Yolov5C3 | 50 | 0.85 | 0.08 | | |
| | Densenet121+TL | 50 | 0.88 | 0.09 | | |
| | in total | 600 | 0.82 | 0.1 | | |

* p<0.05 ** p<0.01

In this study, we examined the performance of multiple cutting-edge target detection models on key performance metrics through careful comparative experimental analyses. Applying the analysis of variance (ANOVA) technique, we observed that the models showed statistically significant differences between each other in both test accuracy (F=52.863, p<0.01) and test loss (F=96.709, p<0.01). In particular, the SE-InceptionV3+L2 model led the way with a high mean value of 0.9 in testing accuracy, demonstrating its remarkable ability to accurately identify positive samples. Similarly, in the loss analysis, the model exhibits a low mean loss value and small standard deviation, reflecting excellent stability during training.

It was also noted that the SE-InceptionV3+L2 model similarly demonstrated excellent performance in the recall ($F=41.885$, $p<0.01$) and F1 score ($F=43.764$, $p<0.01$) analyses, further confirming its ability to effectively identify true positive cases.

These findings emphasize the overall excellence of the SE-InceptionV3+L2 model across a number of key performance metrics including precision, accuracy, recall, and F1 score. The results of this study not only provide practical guidance to relevant researchers in selecting the best model for a particular application but also point the way to future research in hyper-parameter optimization and model architecture tuning.

### *4.6. External dataset validation*

#### *4.6.1. External Datasets and Preprocessing*

To assess the generalisability of the proposed model, an external dataset different from the original dataset was used for validation in this study. This external dataset is the Monkeypox Skin Images Dataset (MSID) [27], which was collected by Diponkor Bala et al. and consists of images from reliable health websites, newspapers, online portals, and public resource sharing. The MSID dataset consists of four categories: monkeypox, chickenpox, measles, and normal skin, where the number of image samples for monkeypox, chickenpox, measles, and normal categories have image sample sizes of 279, 107, 91, and 293, respectively.

Before external dataset validation, the dataset was first pre-processed appropriately. The four categories were reclassified into two categories: monkeypox and non-monkeypox for binary classification studies. Further preprocessing steps included image rotation, size unification, and saturation adjustment to increase the diversity of the dataset. After these processes, the number of images in the monkeypox category increased to 837, while the number of images in the non-monkeypox category was 1,473.

The experiments were conducted in similarly configured hardware environments using the same training parameters as the original dataset experiments to ensure consistency of experimental conditions. This is aimed at evaluating the performance of the model when dealing with different data sources to verify its applicability and accuracy in different scenarios. Fig 19 gives three original images each for monkeypox disease and other categories.

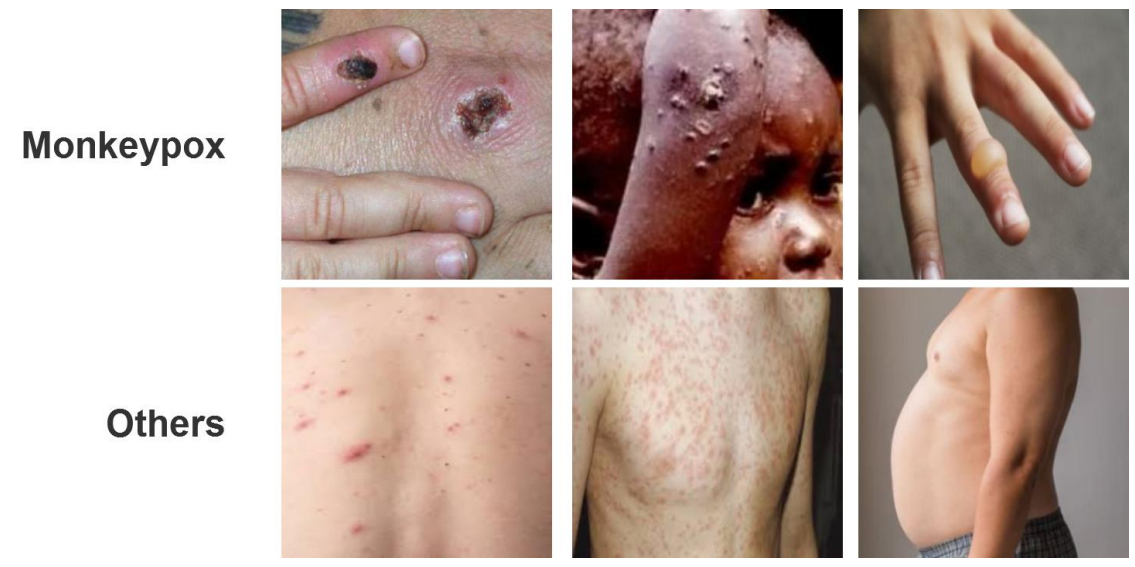

Fig. 19. Example of external dataset.

#### *4.6.2. Performance demonstration of the model on external datasets*

Table 9 demonstrates the performance of the SE-InceptionV3+L2 model compared to other network models on external datasets.

Table 9

Comparison of external dataset modeling results

| Network Models | Test_acc(%) | Test_loss | Test_Precision | Test_Recall | Test_F1 |
|---|---|---|---|---|---|
| SE-InceptionV3+L2 | 95.46 | 0.163 | 0.957 | 0.948 | 0.95 |
| ResNet50V2 | 86.70 | 0.5 | 0.867 | 0.853 | 0.855 |
| VGG-16 | 87.17 | 0.825 | 0.87 | 0.876 | 0.866 |
| InceptionV1 | 92.80 | 0.814 | 0.935 | 0.92 | 0.922 |
| SE-DenseNet | 86.07 | 0.547 | 0.824 | 0.838 | 0.821 |
| ResNeXt-50 | 83.88 | 0.906 | 0.84 | 0.84 | 0.83 |
| ResNet+DenseNet | 88.26 | 0.456 | 0.849 | 0.845 | 0.832 |
| YOLOv5C3 | 91.08 | 0.588 | 0.91 | 0.912 | 0.905 |
| InceptionV3 | 94.21 | 0.189 | 0.944 | 0.944 | 0.937 |
| DenseNet121-TL | 94.98 | 0.17 | 0.956 | 0.956 | 0.952 |

The table clearly shows the highest accuracy of the selected models and their corresponding losses after 50 training epochs on the external dataset. It can be seen that the improved SE-InceptionV3+L2 model achieves 95.46% in terms of accuracy with significantly lower losses than the other models.

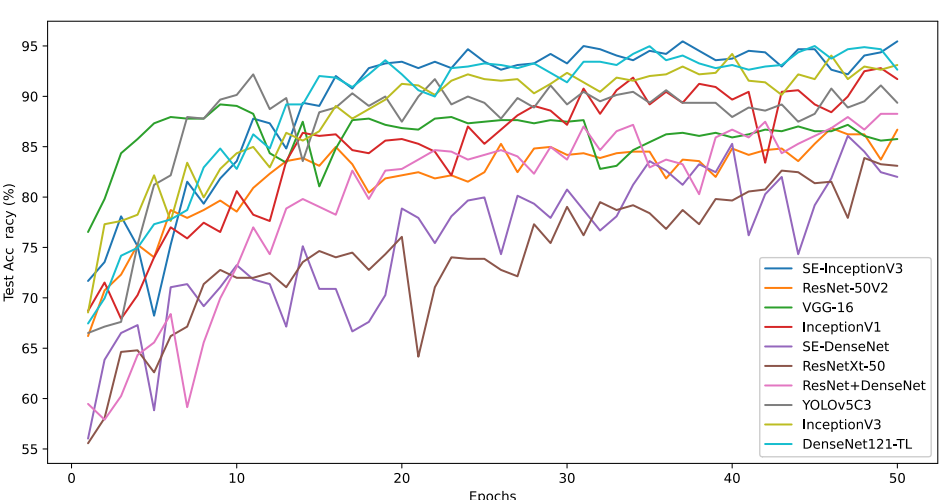

Fig. 20. Comparison of accuracy of each model on external dataset validation set.

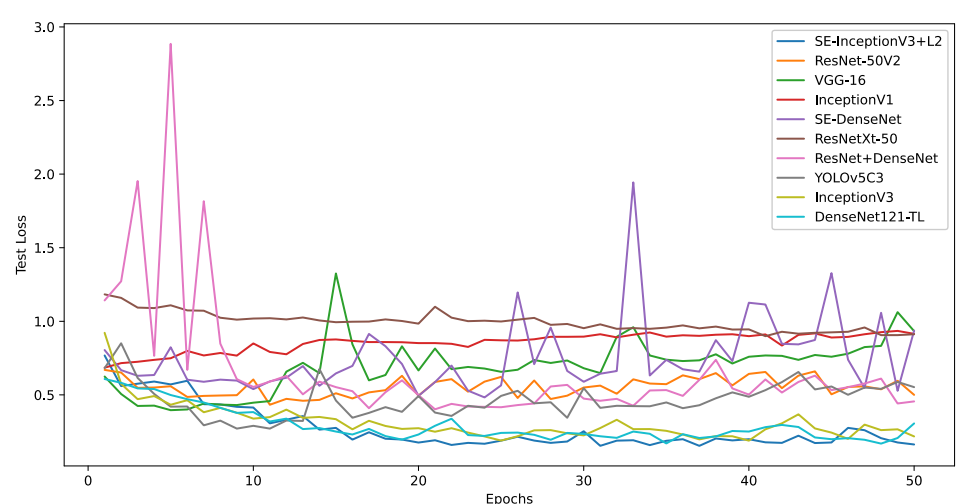

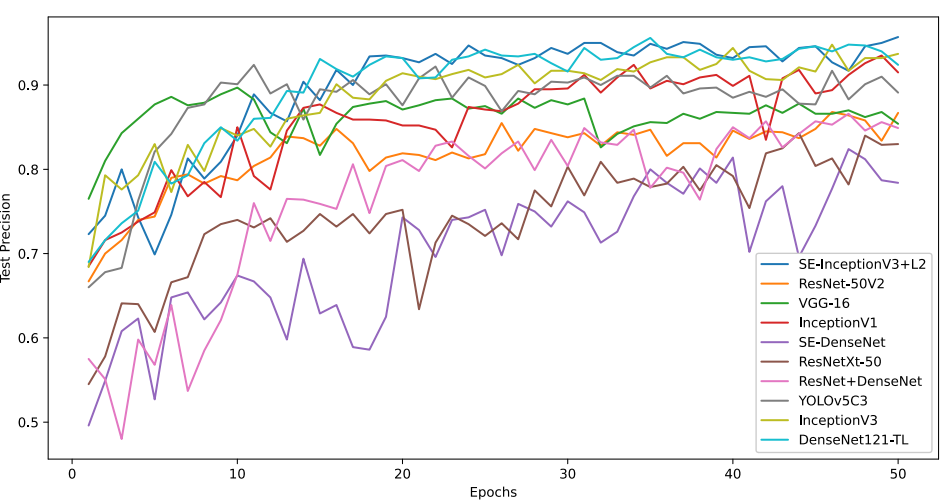

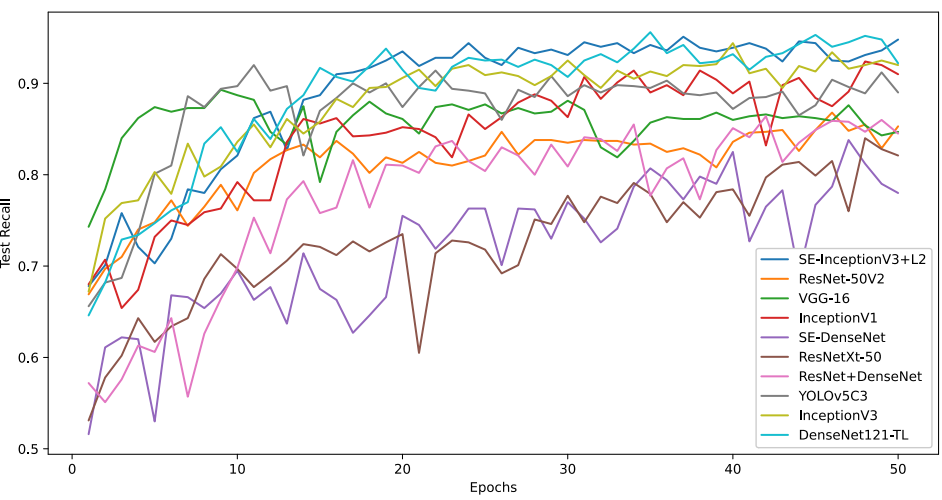

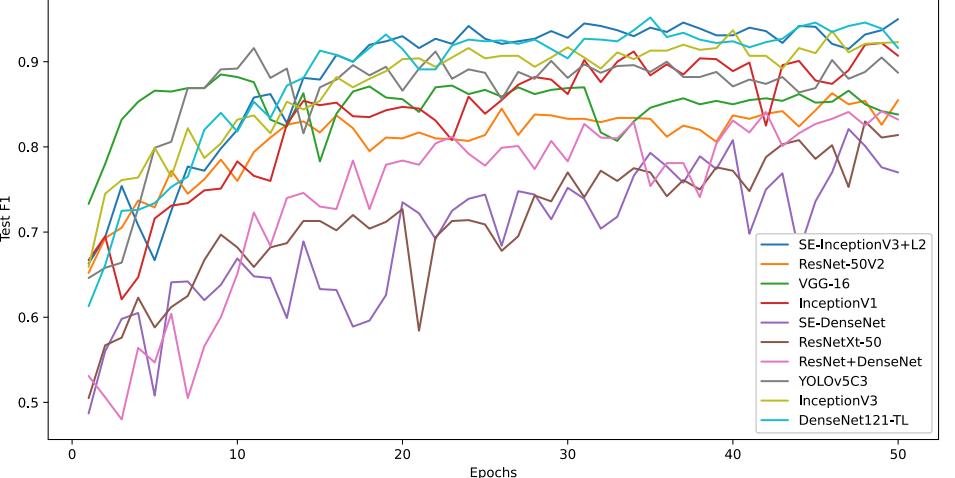

Fig. 21. Comparison of Loss, Precision, Recall, and F1 of each model on external dataset validation set.

*4.6.3. Result analysis and discussion*

As can be seen from the results in Table 9 and Fig 20 and 21, the SE-InceptionV3+L2 model outperforms the other comparative models on the external dataset. This is not only in terms of accuracy but also in terms of loss, precision, recall, and F1 score. These results emphasize the generalisability and effectiveness of the SE-InceptionV3+L2 model when dealing with different data sources.

In summary, the SE-InceptionV3+L2 model demonstrates excellent performance and robustness on datasets from different sources. The model has a broad potential for future applications, especially in the field of different types of medical image recognition.

## 5. CONCLUSION

In order to curb the rapidly spreading monkeypox disease epidemic, an effective method to identify whether a patient is infected with monkeypox disease is desired. In this paper, we propose a network model that improves SE-InceptionV3 and introduces an L2 regularization term in the original InceptionV3 model, which prevents the model from overfitting the training data and improves its generalization ability. In addition, the SE module is added to the model, which can fully utilize the channel information of the model and improve the recognition rate. The network model used in this paper was trained on the monkeypox dataset from the Kaggle platform and was finally able to achieve 96.71% recognition accuracy of monkeypox disease, which is higher than other commonly used deep learning networks. In addition, the model also shows significant advantages in metrics such as loss function, precision, recall, and F1 score. When dealing with challenges specific to monkeypox disease datasets, such as highly similar and complex skin lesions, traditional InceptionV3 models may not work well. Our SE-InceptionV3+L2 model, on the other hand, demonstrates higher accuracy and robustness by improving the focus on key features and enhancing the handling of unknown data. These improvements reflect the significant advantages of our model in distinguishing monkeypox lesions from other similar skin lesions.

However, it is worth noting that our model is relatively complex and requires a large amount of computational resources and data for training. This may be limited in certain resource-limited environments, especially in practical application scenarios such as the medical field. To address this issue, one of the future research directions is to explore model compression techniques to reduce the computational and memory requirements of the model and make it more suitable for resource-constrained environments. In addition, considering data augmentation techniques to generate more training samples and combining them with migration learning is expected to further improve the performance of the model, especially when the size of the dataset is limited. Also, combining trained deep network models with more efficient classifiers is an important research direction to improve the classification accuracy of monkeypox disease. Finally, we recognize the importance of validating models on diverse datasets as part of future research. We plan to keep expanding the range of datasets to test the validity of the model on a wider and diverse range of data.

In summary, despite the remarkable research results, there are still many challenges and opportunities that we need to explore further to continuously improve the application of deep learning techniques in disease detection and other fields. These efforts will hopefully provide more feasible and efficient solutions for medical diagnosis and other fields.

Copyright notice and licence

Author Contributions and Disclaimers

This document is a joint research effort by the listed authors. All listed authors have given full acknowledgement of their contributions to this study and have agreed to the submission and publication of the final version.

Disclaimer of use of third-party content

1. all images, tables and data used in this study that are not original to the authors have been expressly authorised by the original copyright holders and their use is in accordance with arXiv's open access licence requirements.
2. for datasets or images obtained and used from publicly available sources, the source has been clearly acknowledged and it has been ensured that their public use does not violate any copyright agreements.
3. all external links and references included in this article point to publicly available resources, and it has been ensured that the content of these links complies with the relevant open access agreements or licences.

Data and code availability

The datasets and code (where applicable) on which this study relies are shared in accordance with open access principles. Specific details and access methods are given in the appropriate sections of the documentation. For the use of any restricted datasets, this study has followed all applicable usage licensing requirements and the corresponding descriptions and proof of legal use have been provided in the documentation.

Acknowledgements

The Acknowledgements section of this document expresses appreciation only to the individuals or organisations that have directly contributed to or supported the work of this research. Any mention does not imply endorsement or licence of the content or materials used in this document.